\documentclass[letterpaper, 10 pt, conference]{ieeeconf} 

\IEEEoverridecommandlockouts 

\usepackage{cite}
\usepackage{amsmath,amssymb}
\usepackage{graphicx}
\usepackage{booktabs}
\usepackage{tabularx}
\usepackage{array}
\usepackage{multirow}
\usepackage[caption=false,font=footnotesize]{subfig}
\usepackage{url}
\usepackage{float}
\usepackage{xcolor}

\newcolumntype{Y}{>{\raggedright\arraybackslash}X}
\newcolumntype{L}[1]{>{\raggedright\arraybackslash}p{#1}}
\newcolumntype{C}{>{\centering\arraybackslash}p{0.20\linewidth}}

\newcommand{\R}{\mathbb{R}}
\DeclareMathOperator{\LN}{LN}
\DeclareMathOperator{\MLP}{MLP}
\DeclareMathOperator{\softmax}{softmax}
\DeclareMathOperator{\FFN}{FFN}
\DeclareMathOperator{\vecop}{vec}
\DeclareMathOperator{\clamp}{clamp}

\makeatletter
\@ifundefined{IEEEkeywords}{%
}{}
\makeatother

\title{\LARGE \bf
DELTA: Deformable Elevation-Based Local Terrain Attention Encoder for Sparse-Terrain Quadrupedal Locomotion
}

\author{Sanghyun~Park$^{1}$, Moonkyu~Jung$^{1}$, and~Jemin~Hwangbo$^{1}$%
\thanks{$^{1}$The authors are with the Robotics and Artificial Intelligence Lab, Korea Advanced Institute of Science and Technology (KAIST), Yuseong-gu, Daejeon 34141, Republic of Korea (e-mail: shparkv1128@kaist.ac.kr; moonk127@kaist.ac.kr; jhwangbo@kaist.ac.kr).}%
}

\begin{document}

\maketitle
\thispagestyle{empty}
\pagestyle{empty}

\begin{abstract}
Stable quadrupedal locomotion on sparse terrain requires selecting state-relevant terrain evidence for precise foot placement. Model-based foothold planners provide precise foothold selection but rely heavily on explicit model assumptions. Recent attention-based map encoding (AME) studies show that end-to-end reinforcement learning (RL) can learn implicit foothold guidance. However, the computational cost of dense AME encoding grows with map resolution, limiting its scalability to fine-grained sparse terrain. We propose DELTA, a Deformable Elevation-Based Local Terrain Attention encoder. DELTA predicts state-conditioned sampling locations, forms terrain evidence tokens from adaptive local elevation patches, and attends only to a fixed-size token set. With fixed sampling and patch settings, DELTA's encoder cost is independent of map resolution. Experiments show that DELTA achieves final traversal performance comparable to AME at the standard resolution while improving learning efficiency. This fixed encoder cost enables the use of higher-resolution terrain maps, improving traversal on fine-grained sparse terrain. DELTA also demonstrates strong generalization to unseen mixed evaluation courses composed of continuous and discrete terrain elements. Beyond simulation, DELTA demonstrates successful sim-to-real transfer on RAIBO2. Analysis of the learned sampling offsets and attention weights shows that DELTA samples steppable regions and attends to terrain evidence relevant to future touchdowns without foothold labels or attention supervision.
\end{abstract}

\section{Introduction}
\begin{figure}[!t]
\centering
\includegraphics[width=\columnwidth]{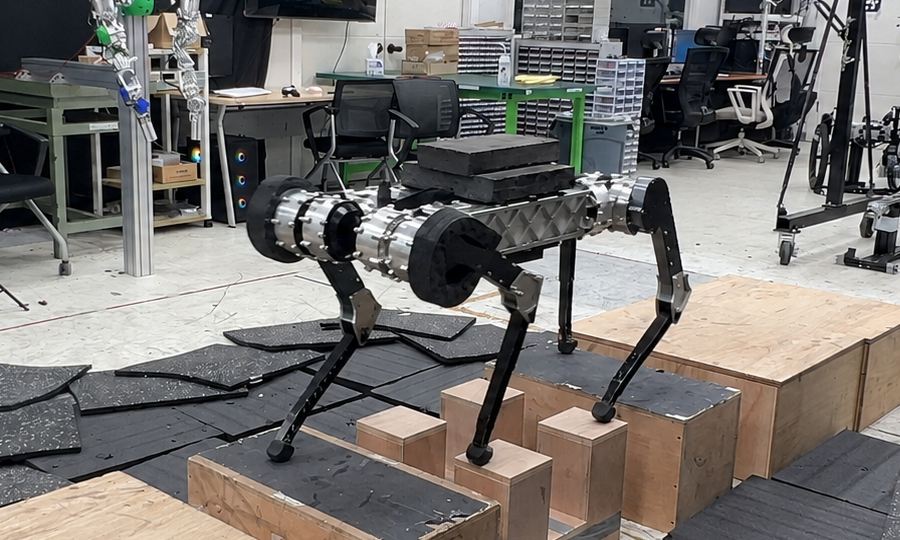}
\caption{Real-robot traversal of an unseen mixed sparse-terrain course using DELTA, requiring precise foot placement on sparse and discontinuous steppable regions without an explicit foothold planner.}
\label{fig:intro}
\end{figure}

Agile and generalizable locomotion is essential for quadrupedal robots operating on complex real-world terrains~\cite{hwangbo2019learning,lee2020learning,miki2022learning,he2025attention,hoeller2024anymal,cheng2024extreme,zhang2024risky}. In sparse terrain such as stepping stones and gaps, only a limited set of regions can support contact, whereas most regions in continuous terrain are steppable. Failing to select a valid foothold or making contact near a gap edge can therefore cause immediate locomotion failure~\cite{he2025attention,jenelten2024dtc,agrawal2022vision,kim2025high}. A sparse-terrain locomotion policy must consequently select and use limited footholds precisely while remaining robust to map noise, state-estimation error, contact uncertainty, and external disturbance. It must also generalize to terrain compositions not encountered during training. Fig.~\ref{fig:intro} shows a representative real-robot sparse-terrain traversal in this work.

Research on sparse-terrain locomotion has traditionally emphasized explicit foothold reasoning. Model-based methods plan feasible foothold sequences and body trajectories from terrain maps, robot dynamics, and contact constraints; representative approaches use trajectory optimization, visual foothold selection with body-pose adaptation, or motion libraries with visual feedback~\cite{jenelten2022tamols,fahmi2023vital,agrawal2022vision}. Hybrid approaches combine learning and model-based components by either using an RL policy to generate foothold plans tracked by a model-based controller~\cite{gangapurwala2022rloc}, or training a policy to track references generated by model-based or optimization-based planners~\cite{jenelten2024dtc,kim2025high}. However, both model-based and hybrid methods remain dependent on explicit model assumptions and online planning. Contact mismatch or unmodeled dynamics can create discrepancies between planned and realized motion, while jointly considering multiple future footholds and body motion increases the cost of online optimization or sampling.

In contrast, reinforcement learning (RL)-based locomotion policies can learn robustness to model errors and disturbances through large-scale simulation and domain randomization. They have also demonstrated sim-to-real transfer across diverse terrains~\cite{hwangbo2019learning,lee2020learning,miki2022learning,rudin2022learning,choi2023learning}. Learning-based perceptive controllers further incorporate exteroceptive observations to address rough and discrete terrains~\cite{miki2022learning,he2025attention,kim2025high}. On sparse terrain, however, robust posture control alone is insufficient: the policy must identify state-relevant terrain evidence for foothold selection. Attention-based map encoding (AME) directly addresses this problem by using proprioception-conditioned attention inside the policy. He et al. showed that the resulting attention can concentrate on steppable regions associated with future footholds without explicit foothold labels~\cite{he2025attention}. Zhang et al. extended this line of work with uncertainty-aware terrain representation and attention-based neural map encoding for agile and generalized locomotion~\cite{zhang2026ame2}. Recent methods further explore selective terrain encoding. GLAD extracts a spatial feature grid from an elevation map and performs state-conditioned sparsification before local attention~\cite{fu2026glad}, whereas TAGA predicts a task-relevant region from visual and proprioceptive cues and crops a local height-scan patch~\cite{li2026taga}. DELTA instead replaces full-map candidate construction with a fixed number of proprioception-conditioned deformable sampling locations and encodes only their local elevation patches before attention.

AME remains the primary comparison because DELTA builds on its state-conditioned attention while replacing full-map candidate construction. The AME encoder first extracts point-wise features over the entire local elevation map using a convolutional neural network (CNN) and then uses all map-cell features as attention candidates~\cite{he2025attention,zhang2026ame2}. Increasing map resolution consequently raises the cost of dense feature extraction, key/value projection, and attention. Because the encoder is evaluated during both rollout collection and policy updates, this cost also increases wall-clock training time. At the same time, finer map resolution is desirable for representing small footholds and narrow terrain features. Dense AME encoding therefore couples terrain precision to computation.

To address this trade-off, we draw inspiration from Deformable DETR (DEtection TRansformer), which reduces dense transformer computation in object detection by attending to a small set of learned sampling points around reference locations~\cite{zhu2021deformable}. We adapt this sparse-sampling principle to terrain encoding and propose DELTA, a Deformable Elevation-Based Local Terrain Attention encoder. DELTA replaces full-map candidate construction with a fixed number of proprioception-conditioned local samples. It predicts deformable sampling offsets from the robot state, refines each location using a scout patch, encodes only the selected elevation patches, and performs multi-head attention over the resulting terrain evidence tokens. Adaptive patch encoding further controls the center--context information represented by the final-layer tokens. The number of processed terrain candidates therefore remains fixed as map resolution increases, decoupling map resolution from encoder candidate count while retaining interpretable attention weights. The main contributions of this paper are as follows.
\begin{itemize}
\item We propose DELTA, a deformable terrain attention encoder whose cost is independent of map resolution for fixed sampling and patch settings. Rather than processing the full map as dense tokens, DELTA samples and encodes a fixed number of local elevation patches conditioned on robot proprioception.
\item We show that DELTA enables resolution-scalable sparse-terrain encoding. At $26\times16$, DELTA preserves AME-level final traversal performance while improving learning and computational efficiency; under the fine-grained curriculum, it improves fine-grained stepping-stone traversal and further benefits from the $41\times25$ map without increasing encoder FLOPs.
\item We show strong generalization to four unseen mixed evaluation courses, where DELTA substantially outperforms AME, and demonstrate successful sim-to-real transfer on the real RAIBO2 platform. Ablation and attention analyses further show that DELTA learns to sample steppable regions and attend to terrain evidence relevant to future touchdowns.
\end{itemize}

\section{Method}
\subsection{Overview}
\begin{figure*}[!t]
\vspace*{4.1pt}
\centering
\includegraphics[width=\textwidth]{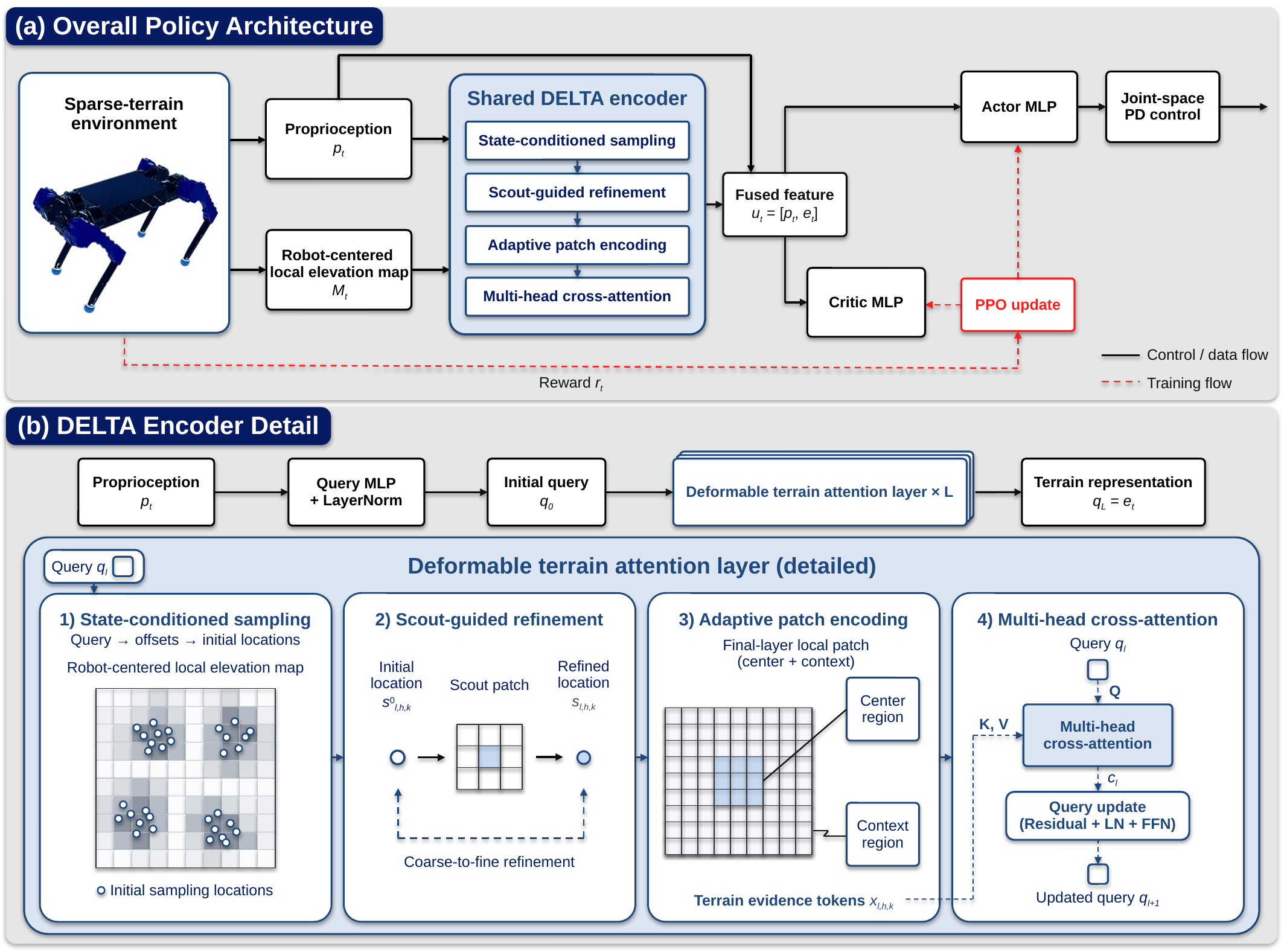}
\caption{Overall policy architecture and DELTA encoder.
(a) DELTA encodes proprioception $p_t$ and the robot-centered local elevation map $M_t$ into a terrain representation $e_t$, which is fused with proprioception and passed to the actor and critic.
(b) DELTA updates the initial query $q_0$ through $L$ deformable terrain attention layers using state-conditioned sampling, scout-guided refinement, adaptive patch encoding, and multi-head cross-attention, yielding $q_L=e_t$.}
\label{fig:architecture}
\end{figure*}

We train an end-to-end actor-critic policy for quadrupedal locomotion on sparse terrain using robot proprioception and a robot-centered local elevation map as inputs. As shown in Fig.~\ref{fig:architecture}, the environment observation before terrain encoding is expressed as
\begin{equation}
o_t = [p_t;\vecop(M_t)] \in \R^{d_p+HWC}.
\end{equation}
Here, $[\cdot;\cdot]$ denotes feature concatenation, $\vecop(\cdot)$ denotes vectorization, and $p_t$ is the proprioceptive observation, consisting of projected gravity, base angular/linear velocity, joint state, previous desired joint-position target, and velocity command. The map $M_t\in\R^{H\times W\times C}$ has height $H$, width $W$, and $C=3$ channels storing base-frame $(x,y,z)$ coordinates. Before entering the policy, the coordinates are normalized as $\tilde{x}=x/1.25$, $\tilde{y}=y/0.75$, and $\tilde{z}=\clamp(z,-0.8,0.8)/0.6$. DELTA generates a compact terrain representation $e_t=f_\theta(p_t,M_t)\in \R^D$. The encoded actor input is then
\begin{equation}
u_t=[p_t;e_t]\in\R^{d_p+D},
\end{equation}
and the policy multilayer perceptron (MLP) predicts the action mean as
\begin{equation}
\mu_t = \MLP_{\pi}(u_t) \in \R^{n_j}.
\end{equation}
During training, $a_t\in\R^{n_j}$ is sampled from the policy with mean $\mu_t$, whereas during evaluation and deployment, $a_t=\mu_t$. The resulting action is converted into desired joint positions as
\begin{equation}
q^{\mathrm{des}}_t = q^{\mathrm{nom}} + \alpha_a a_t, \quad
\tau_t = K_p(q^{\mathrm{des}}_t - q_t) - K_d \dot{q}_t.
\end{equation}
Here, $q_t$, $\dot{q}_t$, and $q^{\mathrm{nom}}$ denote the current joint positions, joint velocities, and nominal joint configuration, respectively. A joint-space proportional--derivative (PD) controller converts the desired joint positions into torque commands, which are clipped to the actuator torque limit $\tau_{\max}$. The encoder and policy are trained end-to-end with proximal policy optimization (PPO)~\cite{schulman2017ppo} and temporal action-mean smoothness regularization, without foothold labels or attention supervision.

\subsection{Deformable Elevation-Based Local Terrain Attention}

DELTA consists of $L$ sequential deformable terrain attention layers, as shown in Fig.~\ref{fig:architecture}(b). The \emph{Query MLP + LayerNorm} block first converts proprioception $p_t$ into the initial query:
\begin{equation}
q_0 = \LN(\MLP_q(p_t)) \in \R^D.
\end{equation}
Here, $\LN$ denotes layer normalization. At each layer, \emph{State-conditioned sampling} and \emph{Scout-guided refinement} determine the sampling locations. Local elevation patches at the refined locations are then encoded into \emph{Terrain evidence tokens}, with \emph{Adaptive patch encoding} used in the final layer. \emph{Multi-head cross-attention} aggregates these tokens using the current query $q_l$, and \emph{Query update (Residual + LN + FFN)} produces $q_{l+1}$. Repeating this sequence through $L$ layers yields the final terrain representation $q_L=e_t$.

\subsubsection{State-Conditioned Sampling}
At layer $l=0,\ldots,L-1$, each attention head $h=1,\ldots,N_h$ uses $K$ sampling points indexed by $k=1,\ldots,K$, with raw reference locations $r_{l,h,k}\in\R^2$. In the first layer, the initial references $r_{0,h,k}$ are initialized once directly in the raw-reference space from a base grid with small random perturbations and thereafter kept fixed. In subsequent layers, the refined reference from the previous layer is passed forward. Query-conditioned initial offsets are computed as
\begin{equation}
\Delta r^0_{l,h,k}=\gamma \tanh(W_s q_l)_{h,k}, \quad
\tilde{r}_{l,h,k}=r_{l,h,k}+\Delta r^0_{l,h,k}.
\end{equation}
Here, $\gamma$ is the offset scale. We bound raw locations to the valid patch-center region using
$b(r)=\boldsymbol{\ell}\odot\tanh(r\oslash\boldsymbol{\ell})$,
where $\odot$ and $\oslash$ denote element-wise multiplication and division, respectively, and $\boldsymbol{\ell}$ contains the per-axis bounds. The bounded location $s^0_{l,h,k}=b(\tilde r_{l,h,k})$ serves as the scout location.

\subsubsection{Scout-Guided Refinement}
A candidate proposed from the query alone may not lie on a locally steppable region. DELTA therefore extracts a $P_s\times P_s$ scout patch around $s^0_{l,h,k}$ by bilinear sampling. Its center-relative elevations are flattened and encoded by a shared scout MLP into a $(D-3)$-dimensional feature, which is concatenated with the normalized three-dimensional base-frame coordinate of the sampled center to form $g_{l,h,k}\in\R^D$. The current query and scout token are concatenated to predict a refinement offset:
\begin{equation}
\Delta r^1_{l,h,k}
=
\gamma\tanh\!\left(\MLP_r([q_l;g_{l,h,k}])\right).
\end{equation}
The refined reference and bounded sampling location are
\begin{equation}
r_{l+1,h,k}
=
\tilde{r}_{l,h,k}+\Delta r^1_{l,h,k}, \quad
s_{l,h,k}=b(r_{l+1,h,k}).
\end{equation}
The bounded location $s_{l,h,k}$ is the \emph{Refined location} in Fig.~\ref{fig:architecture}(b), while $r_{l+1,h,k}$ is passed to the next layer.

\subsubsection{Adaptive Patch Encoding}
Once the refined location $s_{l,h,k}$ is determined, DELTA converts a local elevation patch at that location into a terrain evidence token $x_{l,h,k}\in\R^D$. All patches are extracted by bilinear sampling and represented using center-relative elevations. In the first $L-1$ layers, another $P_s\times P_s$ patch is sampled at the refined location and encoded in the same manner as the scout patch to form the terrain evidence token, whereas the final layer uses a wider adaptive $P\times P$ center--context patch.

A wider final-layer patch provides useful context for edges and gaps but may also mix irrelevant surrounding patterns with the center geometry, creating spurious correlations in unseen terrain compositions. To alleviate this problem, DELTA separates the center-relative $P\times P$ final patch into center $\tilde m_i^c$ and context $\tilde m_i^{\mathrm{ctx}}$ regions. At the final deformable layer, the following operations are independently applied to each sampled point, indexed by $i=(h,k)$:
\begin{equation}
c_i=f_c(\tilde m_i^c), \qquad
u_i=f_{\mathrm{ctx}}(\tilde m_i^{\mathrm{ctx}}),
\end{equation}
\begin{equation}
\eta_i=\sigma\!\left(f_g([c_i;u_i])\right)\in(0,1), \qquad
\phi_i=c_i+\eta_i P_{\mathrm{ctx}}(u_i).
\end{equation}
Here, $f_c$ maps the center to $c_i\in\R^{61}$, and $f_{\mathrm{ctx}}$ maps the context to $u_i\in\R^{61}$; $P_{\mathrm{ctx}}$ projects $u_i$, $f_g$ produces the scalar gate logit, and $\sigma$ denotes the sigmoid function. This allows DELTA to preserve the center geometry while incorporating only the necessary amount of surrounding context.

Finally, the 61-dimensional local terrain feature $\phi_i$ is concatenated with the normalized three-dimensional base-frame coordinates of the sampled center, $\xi_i$, to form the final-layer terrain evidence token $x_{L-1,h,k}$:
\begin{equation}
x_{L-1,h,k}=[\phi_i;\xi_i] \in \R^{61+3}=\R^D.
\end{equation}

\subsubsection{Multi-Head Cross-Attention}
Each deformable terrain attention layer uses single-query multi-head cross-attention. The current query $q_l$ is reshaped into $N_h$ head-specific subvectors $q_{l,h}\in\R^{d_h}$. The terrain evidence tokens $x_{l,h,k}$ are projected into head-specific keys and values, with non-affine root-mean-square (RMS) normalization~\cite{zhang2019rmsnorm} applied to the projected values:
\begin{equation}
\begin{aligned}
k_{l,h,k} &= W_k x_{l,h,k}, \\
v_{l,h,k} &= \operatorname{RMSNorm}(W_v x_{l,h,k}), \\
k_{l,h,k}, v_{l,h,k} &\in \R^{d_h}, \quad d_h = D/N_h.
\end{aligned}
\end{equation}
\begin{equation}
a_{l,h,k}
=
\softmax_k\left(
\frac{q_{l,h}^{\top} k_{l,h,k}}{\sqrt{d_h}}
\right), \quad
c_{l,h}
=
\sum_{k=1}^{K} a_{l,h,k} v_{l,h,k}.
\end{equation}
This value-only normalization prevents a low-attention token from dominating the aggregated context through a large value norm, aligning token contribution magnitudes with the learned attention weights.
The head outputs are concatenated as
$c_l=[c_{l,1};\ldots;c_{l,N_h}]\in\R^D$
and passed through a linear projection. The \emph{Query update (Residual + LN + FFN)} block then produces the query for the next deformable terrain attention layer:
\begin{equation}
\hat{q}_{l+1}
=
\LN(q_l+W_o c_l), \quad
q_{l+1}
=
\LN(\hat{q}_{l+1}+\FFN(\hat{q}_{l+1})).
\end{equation}

Each head in DELTA attends to $K$ sampled candidates, whereas each dense-attention head attends to all $HW$ map-cell candidates. Once the map tensor is available, the costs of patch sampling, local token encoding, key/value projection, attention, and query updates depend on $L$, $N_h$, $K$, $P_s$, $P$, and $D$, rather than on $H$ and $W$. We define these fixed-size operations as the encoder cost; full-map preprocessing and data movement are excluded from this definition.

\subsection{Policy Learning, Reward, and Curriculum}
The terrain encoder, actor, and critic are trained jointly using PPO following the RAIBO2 locomotion framework~\cite{hwangbo2025raibo2}, with temporal action-mean smoothness regularization. Training uses 200 parallel environments with a maximum episode duration of 4.0 s, a simulation frequency of 400 Hz, and a policy control frequency of 100 Hz. During randomized training, the actor and critic share the DELTA encoder parameters, with the actor encoding noisy observations and the critic encoding the corresponding clean observations. Joint-friction and slip randomization are also applied. The policy and DELTA architecture settings are summarized in Table~\ref{tab:training_settings}.

\begin{table}[!t]
\vspace*{4.8pt}
\caption{Main Policy, Control, and DELTA Settings}
\label{tab:training_settings}
\centering
\scriptsize
\renewcommand{\arraystretch}{1.05}
\setlength{\tabcolsep}{0pt}
\resizebox{0.996\columnwidth}{!}{%
\begin{tabular}{@{}l@{\hspace{1.5em}}l@{}}
\toprule
Setting & Value \\
\midrule
$d_p / n_j$
& $48 / 12$ \\
$(H\!\times\!W\!\times\!C)_{\mathrm{std/high}}$
& $26\!\times\!16\!\times\!3$ / $41\!\times\!25\!\times\!3$ \\
Map extent (m)
& $2.0\!\times\!1.2$ \\
$L / D$
& $3 / 64$ \\
$N_h / K$
& $4 / 8$ \\
Patch size $P_s / P$ / center
& $3\!\times\!3$ / $9\!\times\!9$ / $3\!\times\!3$ \\
Patch step (m) / $\gamma$
& $0.08 / 0.25$ \\
$\MLP_q / scout MLP / \MLP_r / \FFN$ hidden layers
& $[64] / [64] / [64] / [256]$ \\
Actor / critic MLP hidden layers
& $[256,128]$ / $[512,256,128]$ \\
$\alpha_a / (K_p,K_d) / \tau_{\max}$
& $0.1 / (100,1) / 71.5$ Nm \\
\bottomrule
\end{tabular}%
}
\end{table}

\begin{figure*}[!t]
\vspace*{4.1pt}
\centering
\includegraphics[width=\textwidth]{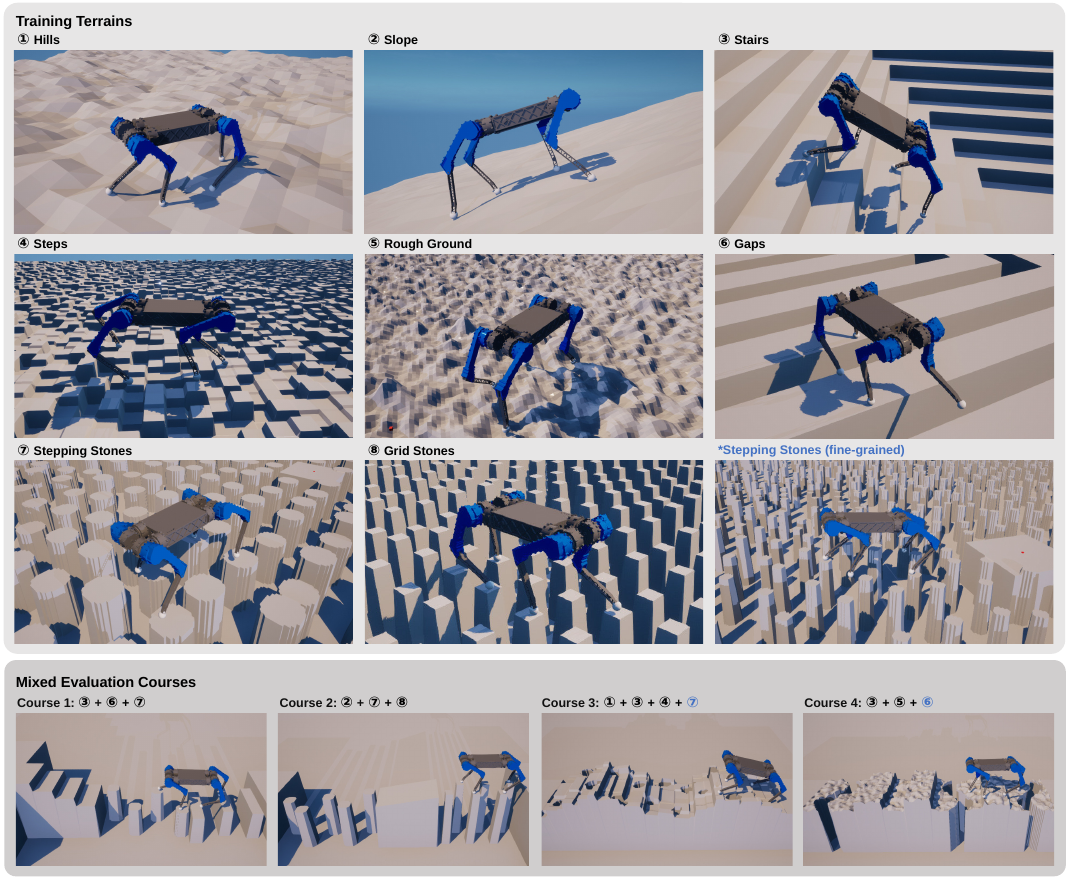}
\caption{Training terrains and mixed evaluation courses. Training includes five continuous terrains (Hills, Slope, Stairs, Steps, Rough Ground) and three discrete terrains (Gaps, Stepping Stones, Grid Stones); the fine-grained stepping stones have half the standard stone radius. The four unseen mixed courses combine: (1) gaps, irregular stepping stones, and ascending stairs; (2) alternating grid stones, a downhill slope, and descending stepping stones; (3) hills, rough tiles, stairs, and terraced platforms with evaluation-only irregular elliptical holes; and (4) rough ground, stairs, and elevated platforms with evaluation-only curved gaps. Blue-highlighted elements in Courses 3 and 4 denote evaluation-only terrain variants.}
\label{fig:terrains}
\end{figure*}

The reward, based on the RAIBO2 locomotion framework~\cite{hwangbo2025raibo2}, is adapted for sparse, discontinuous terrain by relaxing posture regularization, penalizing prolonged stance and undesired contacts more strongly, and modifying foot-related terms for terrain edges, gaps, and limited footholds. Following the positive--negative reward composition of~\cite{ji2022concurrent}, the reward is averaged over the four simulation substeps within each control step and computed as \(r_t=r_t^+\exp(0.1r_t^-)\), where \(r_t^+\) sums the positive reward terms and \(r_t^-\) sums the negative reward terms. A terminal penalty of $-50$ is applied upon failure.

The training set comprises five continuous terrain classes and three discrete terrain classes, as shown in Fig.~\ref{fig:terrains}. Each parallel environment maintains an independent terrain stage from 1 to 10. At episode termination, successful traversal is determined by comparing the expected travel distance with the actual base displacement. The stage is promoted after two consecutive successes and demoted after three consecutive failures; at stage 10, environments are reset to stage 1 with 20\% probability and otherwise resampled from stages 6--10. This design preserves basic locomotion performance while maintaining adaptation to high-difficulty sparse terrains. The stage curriculum is complemented by adaptive terrain sampling, which updates terrain probabilities from running stage and success-rate statistics to emphasize underperforming terrains.

\section{Experiments and Results}
\subsection{Experimental Setup}
We conducted simulation experiments to evaluate DELTA in terms of traversal performance, learning efficiency, computational efficiency, and generalization. The main comparison includes three terrain representation modules. The MLP map encoder encodes the flattened local map into a terrain representation $e_t\in\R^D$. The AME encoder, implemented following He et al.~\cite{he2025attention}, uses all map-cell features as attention candidates. DELTA processes only a fixed number of sampled terrain evidence tokens. All three modules produce terrain representations of the same dimension and use identical downstream actor and critic MLPs.

For the standard-resolution comparison, MLP, AME, and DELTA were trained under identical settings, with only the terrain representation module varied. For the fine-grained comparison, the standard stepping-stone terrain was replaced by its fine-grained variant, and AME and DELTA were each trained at both $26\times16$ and $41\times25$ resolutions over the same physical extent. Simulation training was performed in RaiSim~\cite{hwangbo2018per} using an AMD Ryzen 7 7700 CPU and an NVIDIA RTX 4060 GPU. Each policy was trained for 30,000 iterations, and all final simulation results were evaluated using the corresponding 30,000-iteration checkpoint. Each evaluation comprises 1,000 episodes across five fixed evaluation seeds (200 episodes per seed), with the same seeds used for all compared policies. Real-robot experiments were conducted on the RAIBO2 platform~\cite{hwangbo2025raibo2} using an NVIDIA Jetson AGX Orin.

\subsection{Simulation Traversal Performance and Learning Efficiency}

Simulation traversal performance is evaluated using the stage-10 success rate (SR). For each of the eight training terrains, we fix the terrain at its highest difficulty stage and evaluate with $v_x=1.0$ m/s, $\omega_z^{\mathrm{cmd}}\sim\mathcal{U}(-0.5,0.5)$ rad/s, and a $4.0$-s horizon. An episode is counted as successful if no failure termination occurs and the robot accumulates at least $3.2$ m of body-forward travel. Learning efficiency is quantified by the first evaluation iteration at which a sequence of three consecutive evaluations with a mean stage-10 SR of at least 90\% is completed. Evaluations are conducted at 50-iteration intervals, and the resulting iteration is reported as the mean $\pm$ sample standard deviation over five independent training seeds. Because all methods use the same rollout size, the iteration count is proportional to the number of environment transitions.

Table~\ref{tab:sim} summarizes the three standard-resolution policies. At $26\times16$, AME and DELTA achieved nearly identical overall mean SRs of 96.3\% and 96.4\%, respectively, showing no loss in final traversal performance from DELTA's fixed-size terrain evidence set. DELTA reached the sustained 90\% criterion at iteration $2{,}210\pm160$, requiring 75.1\% fewer iterations than AME and showing lower variability across seeds.

\begin{table}[!t]
\vspace*{4.8pt}
\caption{Simulation Traversal Performance and Learning Efficiency}
\label{tab:sim}
\centering
\scriptsize
\renewcommand{\arraystretch}{1.08}
\setlength{\tabcolsep}{1.3pt}
\resizebox{0.995\columnwidth}{!}{%
\begin{tabular}{@{}l c c c c c c@{}}
\toprule
\multirow{2}{*}{Policy}
& \multicolumn{5}{c}{Success rate (\%) $\uparrow$}
& \multirow{2}{*}{\shortstack{Iterations to\\90\% SR $\downarrow$}} \\
\cmidrule(lr){2-6}
& \shortstack{Continuous\\avg.}
& \shortstack{Stepping\\Stones}
& Gaps
& \shortstack{Grid\\Stones}
& \shortstack{Overall\\avg.}
& \\
\midrule
MLP & 80.8 & 79.9 & 85.4 & 81.2 & 81.3 & N/C \\
AME & \textbf{95.4} & 98.1 & \textbf{99.8} & 95.7 & 96.3
 & $8{,}890 \pm 1{,}475$ \\
DELTA & 94.6 & \textbf{99.5} & 99.7 & \textbf{98.5} & \textbf{96.4}
 & $\mathbf{2{,}210 \pm 160}$ \\
\bottomrule
\end{tabular}%
}
\vspace{0.5ex}

\begin{minipage}{\columnwidth}
\scriptsize
N/C: criterion not reached within 30,000 training iterations.
\end{minipage}
\end{table}

\begin{table}[!t]
\caption{Stepping-Stone SR under the Fine-Grained Curriculum}
\label{tab:fine}
\centering
\scriptsize
\renewcommand{\arraystretch}{1.08}
\setlength{\tabcolsep}{3.0pt}
\resizebox{0.6\columnwidth}{!}{%
\begin{tabular}{@{}l c c c@{}}
\toprule
\multirow{2}{*}{Policy}
& \multirow{2}{*}{Map size}
& \multicolumn{2}{c}{Stepping-stone SR (\%) $\uparrow$} \\
\cmidrule(l){3-4}
& & Standard & Fine-grained \\
\midrule
AME & $26{\times}16$ & 97.9 & 65.7 \\
DELTA & $26{\times}16$ & 98.4 & 84.0 \\
AME & $41{\times}25$ & 92.5 & 68.0 \\
DELTA & $41{\times}25$ & \textbf{99.1} & \textbf{95.9} \\
\bottomrule
\end{tabular}%
}
\end{table}

The fine-grained stepping-stone variant has half the stone radius of the standard variant. All three policies trained with the standard curriculum showed negligible success rates on this variant. Table~\ref{tab:fine} compares AME and DELTA trained with the fine-grained curriculum at $26\times16$ and $41\times25$. The results at $26\times16$ suggest that the approximately $8$-cm grid spacing limits the spatial fidelity of the terrain representation for the fine-grained terrain, whereas $41\times25$ reduces the spacing to $5$ cm and provides a finer representation. Nevertheless, at the same $26\times16$ resolution, DELTA achieved 84.0\% fine-grained SR compared with 65.7\% for AME. At $41\times25$, DELTA reached 95.9\%, whereas AME reached only 68.0\% within the same 30,000-iteration training budget, indicating that increasing map resolution alone was insufficient for dense AME to achieve comparable fine-grained traversal performance.

\subsection{Computational Efficiency}
Computational efficiency is evaluated using encoder floating-point operations (FLOPs) and mean training iteration time. Encoder FLOPs are calculated per forward pass using two operations per multiply--accumulate, excluding nonlinearities, normalization, and bilinear sampling. Wall-clock training efficiency is measured as the mean time per training iteration under identical settings, excluding iterations that perform evaluation or checkpointing. Relative training speed is reported relative to AME at $26\times16$, which is set to 1.00.

\begin{table}[!t]
\vspace*{4.8pt}
\caption{Encoder FLOPs and Wall-Clock Training Efficiency}
\label{tab:efficiency}
\centering
\scriptsize
\renewcommand{\arraystretch}{1.06}
\setlength{\tabcolsep}{1.4pt}
\resizebox{0.9\columnwidth}{!}{%
{
\begin{tabular}{@{}l c c c c@{}}
\toprule
Encoder setting
& Map size
& \shortstack{Encoder\\FLOPs (M) $\downarrow$}
& \shortstack{Mean iter.\\time (s) $\downarrow$}
& \shortstack{Relative training\\speed ($\times$) $\uparrow$} \\
\midrule
AME
& $26{\times}16$ & 27.570 & 20.262 & 1.00 \\
DELTA
& $26{\times}16$ & 4.496 & 8.239 & 2.46 \\
\midrule
AME
& $41{\times}25$ & 67.910 & 46.558 & 0.44 \\
DELTA
& $41{\times}25$ & 4.496 & 10.541 & 1.92 \\
\bottomrule
\end{tabular}}%
}
\end{table}

Table~\ref{tab:efficiency} shows that DELTA maintains 4.496 M encoder FLOPs at both resolutions, whereas AME increases from 27.570 M at $26\times16$ to 67.910 M at $41\times25$ as its candidate count grows from 416 to 1,025. DELTA therefore reduces encoder FLOPs by 83.7\% at the standard resolution and 93.4\% at the higher resolution. Mean iteration times correspond to $2.46\times$ and $4.42\times$ speedups at $26\times16$ and $41\times25$, respectively. Thus, DELTA's computational advantage widens with map resolution while its encoder FLOPs remain fixed. Across the same five training seeds, DELTA required $5.06\pm0.33$ h of training time to reach the sustained 90\% mean stage-10 SR criterion, excluding evaluation and checkpointing, corresponding to a $9.9\times$ speedup over AME.

\subsection{Generalization to Mixed Evaluation Courses}

Generalization performance was evaluated using the three $26\times16$ policies trained with the standard curriculum on four unseen mixed evaluation courses that were not used during training. Each course was evaluated at $v_x=1.0$ m/s, and an episode was counted as successful if the robot completed the course within its course-specific time limit. The evaluation metrics were success rate (SR) and mean progress, with the average SR and progress computed across the four courses.

Table~\ref{tab:generalization} reports the course-wise results. DELTA achieved an average SR of 96.6\% and average progress of 98.4\%, compared with 26.0\% and 62.7\% for AME, respectively. The 70.6-percentage-point SR gap contrasts sharply with their nearly identical performance on the individually evaluated training terrains, indicating that AME's degradation emerges under the unseen terrain compositions. Qualitative inspection of AME rollouts and attention maps showed failures at transitions between terrain types and in sections where continuous and discrete elements co-occurred in the local map; in these cases, attention did not consistently shift toward upcoming foothold-relevant regions, and locomotion concurrently became unstable. In these failure cases, the local map simultaneously contains the outgoing terrain, terrain boundaries, upcoming footholds, and surrounding terrain that is not directly relevant to the next contact. Because AME first constructs dense CNN features over the full map and then relies on proprioception-conditioned attention to select among all candidates, these heterogeneous cues remain in the candidate set, requiring the attention mechanism to suppress evidence from the outgoing or locally irrelevant terrain while shifting toward sparse upcoming footholds. This provides a plausible explanation for the observed attention and locomotion degradation. The degradation was particularly pronounced in Courses 3 and 4, which contain evaluation-only terrain geometries. DELTA reduces this ambiguity by refining candidate locations using state and local geometry before attention and adaptively controlling the surrounding context, consistent with the ablation results below.

\begin{table}[!t]
\vspace*{4.8pt}
\caption{Generalization Performance on Four Unseen Mixed Evaluation Courses}
\label{tab:generalization}
\centering
\scriptsize
\renewcommand{\arraystretch}{1.05}
\setlength{\tabcolsep}{1.4pt}
\resizebox{0.996\columnwidth}{!}{%
\begin{tabular}{@{}l cc cc cc@{}}
\toprule
Course
& \multicolumn{2}{c}{MLP}
& \multicolumn{2}{c}{AME}
& \multicolumn{2}{c}{DELTA} \\
\cmidrule(lr){2-3}\cmidrule(lr){4-5}\cmidrule(l){6-7}
& SR (\%) $\uparrow$
& \shortstack{Mean\\progress (\%) $\uparrow$}
& SR (\%) $\uparrow$
& \shortstack{Mean\\progress (\%) $\uparrow$}
& SR (\%) $\uparrow$
& \shortstack{Mean\\progress (\%) $\uparrow$} \\
\midrule
Course 1 & 0.0 & 75.3 & 41.8 & 62.9 & \textbf{97.6} & \textbf{98.5} \\
Course 2 & 12.6 & 65.4 & 53.8 & 69.0 & \textbf{99.3} & \textbf{99.6} \\
Course 3 & 0.0 & 47.5 & 6.2 & 59.9 & \textbf{97.4} & \textbf{98.0} \\
Course 4 & 1.4 & 57.2 & 2.2 & 58.9 & \textbf{92.0} & \textbf{97.5} \\
\midrule
Average
& 3.5 & 61.4
& 26.0 & 62.7
& \textbf{96.6} & \textbf{98.4} \\
\end{tabular}%
}
\end{table}

\subsection{Ablation and Attention Analysis}

Table~\ref{tab:ablation} analyzes the contributions of Scout-guided refinement and Adaptive patch encoding. All variants use the $26\times16$ map and the standard curriculum. The ablated variants remove scout-patch-based sampling refinement, center--context separation, or deformable sampling. In the Fixed Sampling variant, both query-conditioned and scout-based offsets are removed, so all layers use the initialized base-grid reference locations while all other components remain unchanged. Traversal performance is evaluated using the stage-10 stepping-stone SR and the average SR across the four mixed evaluation courses. On stage-10 stepping stones, the Ground-Truth (GT) Steppable-Surface Ratio measures the percentage of final-layer sampling points located on steppable surfaces. Future-Touchdown (FT) Attention Ratio measures, for each episode, the head-averaged normalized attention at foot touchdown locations, averaged over the preceding 100\,ms and across all touchdowns, relative to the uniform baseline of $1/32$; $1\times$ corresponds to uniform attention. The last two metrics are reported as episode means $\pm$ standard deviations.

\begin{table}[!t]
\caption{Ablation of DELTA Components}
\label{tab:ablation}
\centering
\scriptsize
\renewcommand{\arraystretch}{1.02}
\setlength{\tabcolsep}{1.5pt}
\begin{tabularx}{\columnwidth}{@{}L{0.19\columnwidth} *{4}{>{\centering\arraybackslash}X}@{}}
\toprule
\multirow{2}{*}{Variant}
& \multicolumn{2}{c}{Traversal (\%) $\uparrow$}
& \multicolumn{2}{c}{Sampling \& attention $\uparrow$} \\
\cmidrule(lr){2-3}\cmidrule(l){4-5}
& \shortstack{Stepping-Stone\\SR}
& \shortstack{Mixed-Course\\Avg. SR}
& \shortstack{GT steppable\\surface ratio (\%)}
& \shortstack{FT attn.\\ratio ($\times$)} \\
\midrule
DELTA
& \textbf{99.5} & \textbf{96.6}
& $\mathbf{85.4\!\pm\!1.0}$
& $\mathbf{2.11\!\pm\!0.16}$ \\
\mbox{w/o Scout}
& 95.2 & 75.0
& $68.7\!\pm\!1.8$
& $1.33\!\pm\!0.07$ \\
\mbox{w/o Adapt. Patch}
& 97.3 & 56.1
& $75.4\!\pm\!2.8$
& $1.84\!\pm\!0.10$ \\
\mbox{Fixed Sampling}
& 96.0 & 28.2
& $31.6\!\pm\!6.2$
& $1.18\!\pm\!0.24$ \\
\bottomrule
\end{tabularx}
\end{table}

DELTA placed 85.4\% of its final-layer samples on steppable surfaces and assigned $2.11\times$ the uniform attention at subsequent touchdown locations. Removing Scout-guided refinement reduced the GT Steppable-Surface Ratio from 85.4\% to 68.7\%, indicating that scout-based refinement helps move sampling locations toward steppable regions. In contrast, removing Adaptive patch encoding substantially reduced the mixed-course average SR from 96.6\% to 56.1\%, while the stepping-stone SR remained high at 97.3\%. This result suggests that Adaptive patch encoding contributes primarily to generalization by controlling the center--context information represented by the sampled terrain evidence. Fixed Sampling yielded the lowest mixed-course SR and the weakest sampling and attention metrics, indicating the importance of deformable sampling.

\begin{figure}[!t]
\vspace*{4.1pt}
\centering
\includegraphics[width=\columnwidth]{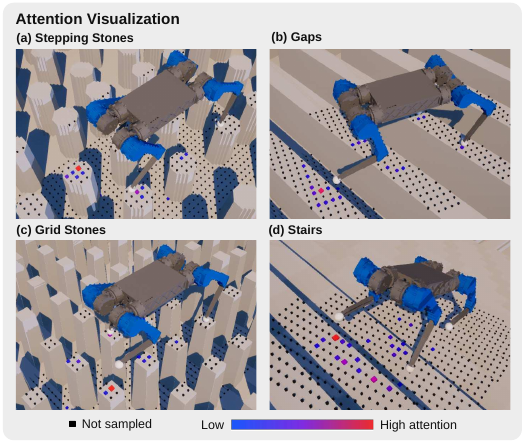}
\caption{Final-layer sampling locations and head-averaged attention of DELTA at $41\times25$ resolution. Attention weights are bilinearly distributed over the elevation map and averaged across heads; black indicates zero weight and blue-to-red indicates increasing attention.}
\label{fig:attention_analysis}
\end{figure}

Fig.~\ref{fig:attention_analysis} qualitatively supports these results. Although DELTA is trained without foothold labels or attention supervision, final-layer samples tend to lie on steppable regions, while the attention maps emphasize sampled terrain evidence associated with subsequent touchdown locations.

\subsection{Real-Robot Experiments}

To demonstrate sim-to-real transfer, we directly deployed the high-resolution DELTA policy trained in simulation on the real RAIBO2 platform~\cite{hwangbo2025raibo2}, without additional real-world training or policy adaptation. The policy was deployed without an additional foothold planner or online trajectory optimization. For hardware safety, real-robot deployment was limited to the high-resolution DELTA policy; the low-resolution DELTA policy and all non-DELTA baselines were evaluated only in simulation.

Intel RealSense D430 depth cameras were mounted at the front and rear of the robot, and all terrain observations were generated solely from their measurements. A robot-centered local elevation map covering a $2.0\times1.2$ m region was constructed using a $41\times25$ grid. The TensorRT policy ran at 100 Hz on the Jetson AGX Orin. Except for filling NaN values and removing sensor spikes, no learned map reconstruction, inpainting, or additional terrain filtering was applied.

\begin{figure}[!t]
\vspace*{4.1pt}
\centering
\includegraphics[width=\columnwidth]{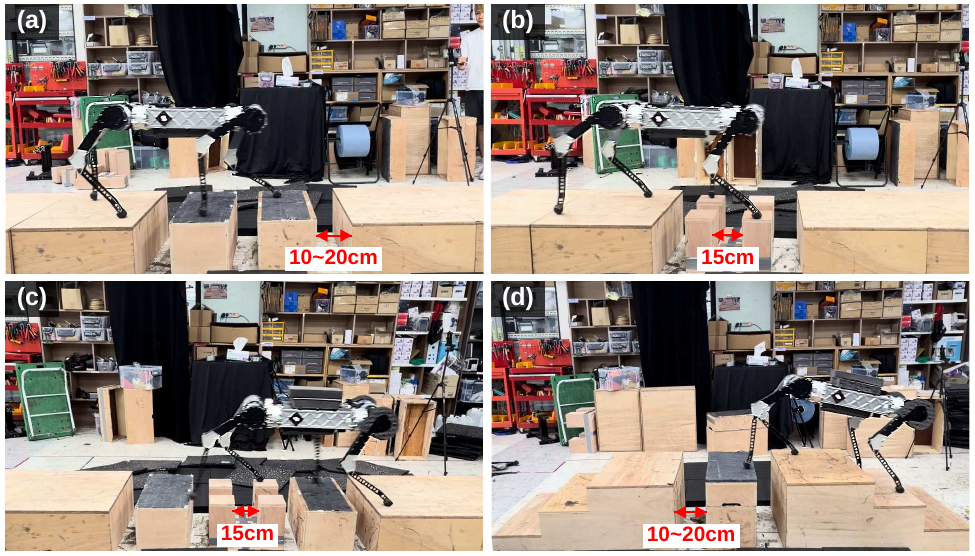}
\caption{Representative real-world courses: (a) Gaps, (b) Stepping Stones, (c) Gaps--Stepping Stones--Gaps, and (d) Stairs--Gaps--Stairs. Gap distances are 10--20 cm, with 15-cm stepping-stone spacing.}
\label{fig:real_robot}
\end{figure}

High-resolution DELTA was evaluated on seven real-world terrain courses: three basic courses consisting of two gaps, three gaps, and $15\,\mathrm{cm} \times 15\,\mathrm{cm}$ stepping stones, and four composite courses consisting of Gaps--Stepping Stones, Gaps--Stepping Stones--Gaps, Stairs--Gaps--Stairs, and Stairs--Two Gaps--Stairs. Representative real-world courses are shown in Fig.~\ref{fig:real_robot}. DELTA successfully completed all 21 trials (three per course; 100\% SR). These results demonstrate successful sim-to-real transfer using real sensor measurements with only minimal map preprocessing.

\section{Conclusion}

We proposed DELTA, a Deformable Elevation-Based Local Terrain Attention encoder that uses proprioception-conditioned deformable sampling to maintain a fixed encoder cost across map resolutions. DELTA matched AME's final traversal performance at the standard resolution while requiring 75.1\% fewer iterations to reach the sustained 90\% SR criterion, yielding a $9.9\times$ wall-clock speedup. At high resolution, DELTA achieved 95.9\% SR on fine-grained stepping stones with 93.4\% fewer encoder FLOPs than AME and improved mixed-course average SR by 70.6 percentage points. Ablation and attention analyses showed that Scout-guided refinement improves steppable-region sampling, while Adaptive patch encoding contributes to generalization.

In real-robot experiments, the high-resolution DELTA policy transferred directly from simulation to the RAIBO2 platform without real-world training or policy adaptation. It completed all 21 trials across seven courses using real sensor measurements, without an explicit foothold planner or online trajectory optimization. Future work will extend DELTA to uncertainty-aware and multi-channel terrain representations incorporating additional cues such as semantics and friction.

\end{document}